\documentclass{article}
\PassOptionsToPackage{numbers}{natbib}
\usepackage[preprint]{neurips_2025}

\usepackage[utf8]{inputenc} 
\usepackage[T1]{fontenc}    
\usepackage{hyperref}       
\usepackage{url}
\usepackage{graphicx}
\usepackage{algorithm}
\usepackage{algorithmic}
\usepackage{booktabs}       
\usepackage{amsfonts}       
\usepackage{nicefrac}       
\usepackage{microtype}      
\usepackage{xcolor}         
\usepackage[numbers]{natbib}
\usepackage{graphicx}       
\usepackage{subcaption}     
\usepackage{amsmath} 
\usepackage{multirow}
\usepackage{multicol}

\title{\textit{SatelliteFormula}: Multi-Modal Symbolic Regression from Remote Sensing Imagery for Physics Discovery}

\author{%
  Zhenyu Yu \\
  Universiti Malaya \\
  \texttt{yuzhenyuyxl@foxmail.com} \\
  \And
  Mohd. Yamani Idna Idris \\
  Universiti Malaya \\
  \And
  Pei Wang \\
  Kunming University of \\Science and Technology \\
  \And
  Yuelong Xia \\
  Yunnan Normal University \\
  \And
  Fei Ma \\
  Guangming Laboratory \\
  \And
  Rizwan Qureshi \\
  University of Central Florida \\
}

\begin{document}

\maketitle

\begin{figure*}[h]
    \centering
    \includegraphics[width=1.0\linewidth]{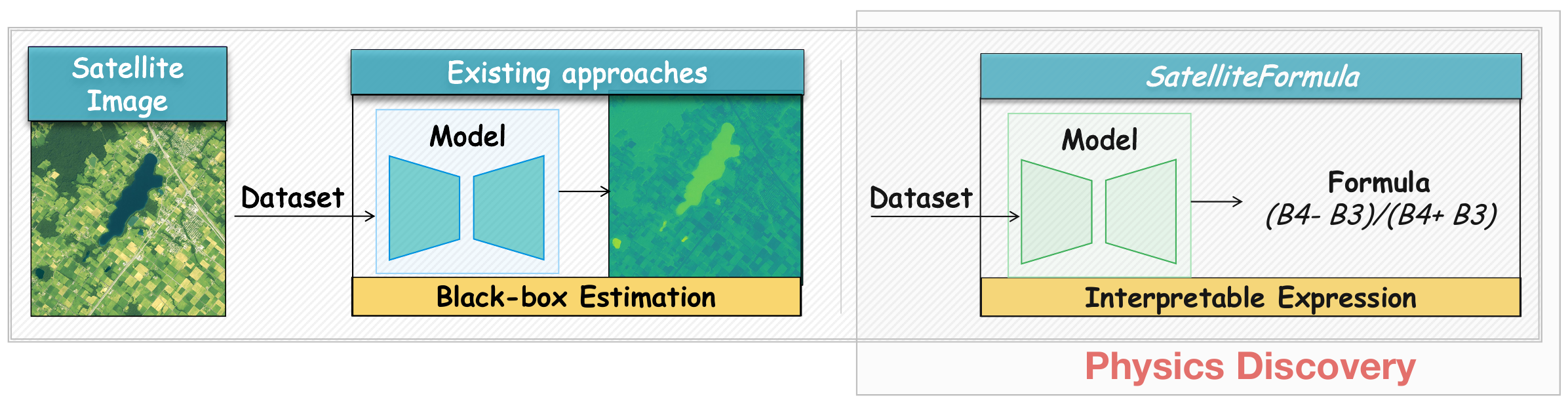} 
    \caption{Motivation. Traditional methods rely on black-box models for estimation, lacking interpretability and physical insights. In contrast, \textit{SatelliteFormula} directly infers symbolic expressions from satellite imagery, enabling physics discovery.}
    \label{fig_motivation}
\end{figure*}

\begin{abstract}
We propose \textit{SatelliteFormula}, a novel symbolic regression framework that derives physically interpretable expressions directly from multi-spectral remote sensing imagery. Unlike traditional empirical indices or black-box learning models, \textit{SatelliteFormula} combines a Vision Transformer-based encoder for spatial-spectral feature extraction with physics-guided constraints to ensure consistency and interpretability. Existing symbolic regression methods struggle with the high-dimensional complexity of multi-spectral data; our method addresses this by integrating transformer representations into a symbolic optimizer that balances accuracy and physical plausibility. Extensive experiments on benchmark datasets and remote sensing tasks demonstrate superior performance, stability, and generalization compared to state-of-the-art baselines. \textit{SatelliteFormula} enables interpretable modeling of complex environmental variables, bridging the gap between data-driven learning and physical understanding. 
\end{abstract}


\section{Introduction}
\label{sec:intro}
\textbf{Remote sensing imagery} provides rich multi-spectral data critical for assessing vegetation health, soil moisture, and surface temperature~\cite{cheng2016survey, zahra2024current,tan2023spatiotemporal}. While traditional indices (e.g., NDVI, SAVI) rely on expert-designed formulas~\cite{somvanshi2020comparative,yu2025satellitecalculator,yu2025qrs}, and deep learning models (e.g., Transformers, UNet, GNNs) offer strong predictive power~\cite{hohl2024opening,yu2025cloud}, both approaches fall short in deriving explicit, physically interpretable expressions. Symbolic Regression (SR) has emerged as a compelling alternative, capable of uncovering latent physical laws~\cite{li2024advancing,yu2025prompt}. However, applying SR to remote sensing is hindered by the high-dimensional and spatial-spectral complexity of satellite data~\cite{kucklick2023tackling, wang2023comprehensive}. The central \textbf{problem} remains: how can we directly infer physically grounded expressions from raw imagery? Solving this would significantly enhance environmental monitoring and scientific interpretability in geospatial analysis~\cite{Zhang2023}.

\textbf{Symbolic regression (SR)} has progressed from early Genetic Programming (GP) approaches—capable of discovering closed-form expressions but limited by computational inefficiency in high-dimensional settings~\cite{Koza2020,yang2020computer,yu2024capan}—to physics-informed frameworks like AI Feynman and PINNs, which integrate domain constraints for improved interpretability but remain restricted to tabular data~\cite{Udrescu2020, Cranmer2022,yu2025rainy,yu2025point}. Recent Transformer-based models (e.g., SymbolicGPT, NeSymReS, TPSR) have enabled scalable and efficient expression discovery through sequence modeling of symbolic forms~\cite{Kamienny2022, Biggio2023, wang2024multi}. Building on these advances, Multi-Modal Symbolic Regression (MMSR) has emerged to address the broader challenge of extracting interpretable expressions from multi-modal inputs~\cite{Liu2024,yu2025diffusion}.

\textbf{Challenges.} Despite MMSR's ability to handle multi-modal data, its data encoder remains optimized for numerical tabular data, making it challenging to effectively capture the complex spatial and spectral features present in remote sensing imagery~\cite{sapkota2024multi, raza2025responsible}. Multi-spectral data often exhibit high spatial dependencies and spectral variance, which traditional encoding mechanisms struggle to learn \cite{Wang2023,yu2025cloud}. This limitation not only affects the accuracy of physical modeling but also hinders the applicability to more advanced geospatial tasks \cite{Li2024}.

\textbf{Our Approach.} To overcome the limitations of prior SR methods on remote sensing data, we introduce \textit{SatelliteFormula}, a symbolic regression framework tailored for multi-spectral imagery. Unlike existing approaches reliant on tabular data or handcrafted features, \textit{SatelliteFormula} directly operates on image inputs via a dedicated image encoder that captures fine-grained spatial and spectral patterns critical for physics-aware modeling.

\textbf{Key Contributions.}
\begin{itemize}
\item We propose \textit{SatelliteFormula}, the first symbolic regression framework to derive physically interpretable expressions directly from multi-spectral remote sensing imagery.
\item We design a multi-spectral image encoder that preserves both spatial structure and spectral variance for symbolic inference.
\item We introduce physics-based constraints via a physical loss, enabling selection of expressions consistent with known physical principles.
\end{itemize}

\section{Related Work}
\subsection{Traditional Symbolic Regression Methods}

Symbolic regression (SR), based in Genetic Programming (GP), has long been used for interpretable scientific discovery~\citep{koza1994genetic}. GP evolves mathematical expressions to fit data but suffers from high computational cost, limited scalability, and a tendency toward overly complex or suboptimal expressions~\citep{LaCava2021}. While its interpretability is appealing, these limitations hinder its effectiveness on modern, high-dimensional datasets. Recent efforts have improved scalability by integrating meta-learning and neural-guided search~\citep{Petersen2023}. To overcome GP’s limitations, physics-guided symbolic regression methods such as AI Feynman~\citep{udrescu2020ai} and SINDy~\citep{brunton2016discovering} integrate domain knowledge to constrain the search space and enhance interpretability. AI Feynman applies dimensional analysis, while SINDy identifies sparse representations of dynamical systems. Although effective for tabular data, these methods lack support for complex multi-modal inputs like remote sensing imagery. Hybrid approaches combining physics-informed neural networks (PINNs) with symbolic regression aim to address this gap, but still struggle with spatial and spectral dependencies inherent in high-dimensional image data~\citep{Wang2024}.

\subsection{Transformer-based Symbolic Regression}

Transformer architectures have redefined symbolic regression by framing expression generation as a sequence modeling task, offering greater scalability and efficiency than GP-based approaches~\citep{Vaswani2017}. For example, SymbolicGPT~\citep{kamienny2021symbolicgpt} treats symbolic expression generation as a language modeling problem, generating token-by-token expressions to capture mathematical structure. Although, it outperforms traditional SR methods in accuracy and complexity~\citep{valipour2021symbolicgpt}, it incurs high training costs for large input spaces. Recent variants incorporate sparse attention to reduce overhead~\citep{Zhang2024}. TPSR~\citep{landajuela2023tpsr} combines Transformer decoding with Monte Carlo Tree Search (MCTS) to efficiently explore expression trees and prune redundancies. However, its architecture remains limited in handling heterogeneous data modalities~\citep{Chen2023}.

NeSymReS~\citep{biggio2021nesymres} achieves strong generalization through pretraining on large symbolic corpora. While effective on tabular benchmarks, its original design is not optimized for spatial or multi-modal data. Recent extensions explore multi-task learning to broaden its applicability\citep{Liu2024}. Despite these advances, Transformer-based symbolic regression methods have not fully explored applications to multi-modal data such as remote sensing imagery, underscoring the need for a novel framework capable of handling spatial and spectral dependencies~\citep{Wang2023}.

\subsection{Symbolic Regression for Remote Sensing}

Remote sensing analysis has traditionally relied on handcrafted empirical indices such as NDVI and SAVI \citep{huete2018vegetation}, which are tailored to specific spectral bands and exhibit limited generalizability. Black-box models like Random Forest and XGBoost \citep{chen2016xgboost} excel in predictive tasks but lack interpretability and the ability to uncover physical laws. Similarly, convolutional neural networks\citep{zhu2017deep} and probabilistic programming approaches\citep{goodman2012probabilistic} have been explored for remote sensing but struggle to balance interpretability with physical constraints. Recent studies have highlighted the need for hybrid models that integrate domain knowledge to improve both accuracy and interpretability in remote sensing applications~\citep{Zhang2023, Li2024}.

To date, symbolic regression has not been systematically applied to remote sensing tasks. Existing SR methods—e.g., AI Feynman and NeSymReS—are tailored for tabular data and lack mechanisms to model the spatial-spectral dependencies present in multi-spectral imagery~\citep{Udrescu2020, Biggio2021}. While remote sensing research has explored hybrid models that integrate machine learning with physical priors (e.g., neural radiative transfer models~\citep{verrelst2019neural}), these models prioritize predictive accuracy over interpretability. Transformer-based architectures have been adopted for remote sensing~\citep{Yuan2022}, but often neglect physical constraints, limiting scientific insights. We propose~\textbf{\textit{SatelliteFormula}}, the first symbolic regression framework for remote sensing, combining a Swin Transformer encoder~\citep{Liu2021} to capture spatial-spectral structure with physics-based constraints to ensure interpretable, physically consistent expressions. This approach addresses both the generalization gap of traditional SR and the opacity of black-box models, enabling interpretable scientific discovery in environmental monitoring and climate analysis~\citep{Wang2024}.
\section{Method}
\subsection{Overview}

\begin{figure}[ht]
    \centering
    \includegraphics[width=1.0\linewidth]{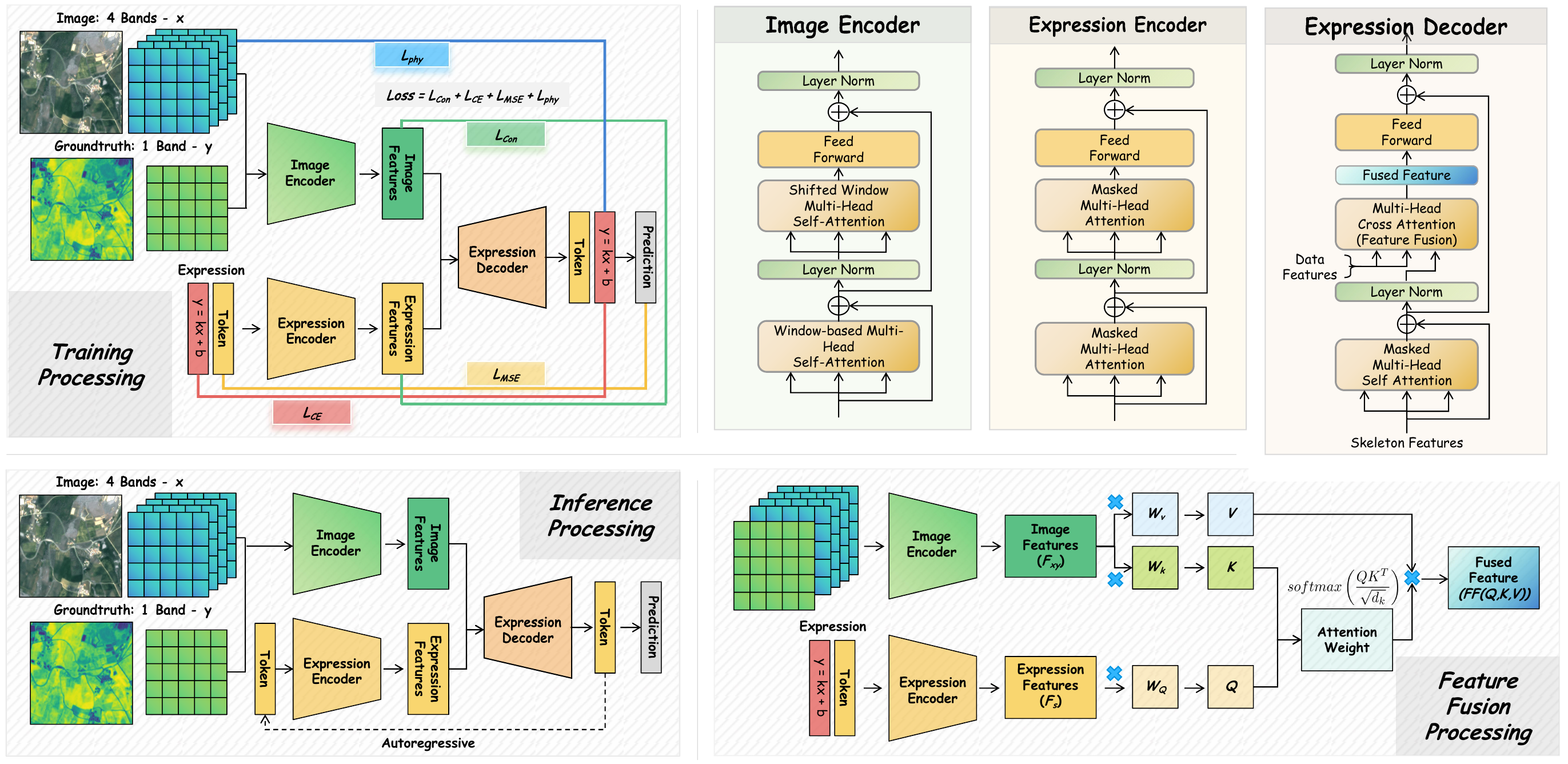}
    \caption{Overview of the \textit{SatelliteFormula} framework. The \textbf{training stage} extracts multi-scale spatial-spectral features and maps them to symbolic expressions under physics constraints. The \textbf{inference stage} applies the trained model to new imagery for interpretable expression generation. \textbf{Module details} include the Image Encoder, Expression Encoder, and Decoder, with \textbf{Feature Fusion} integrating spatial-spectral features for symbolic output.}
    \label{fig_overview}
\end{figure}

\textit{SatelliteFormula} is a symbolic regression framework tailored for remote sensing imagery, designed to infer interpretable mathematical expressions that reflect underlying physical processes. As shown in Figure~\ref{fig_overview}, the framework operates in two stages: \textbf{training} and \textbf{inference}. During training, the model learns to map multi-spectral image features to symbolic expressions using paired data. In inference, it generates expressions for unseen imagery, enabling downstream physical interpretation and scientific analysis.

\subsection{Problem Definition}

The goal of symbolic regression in remote sensing is to discover explicit mathematical expressions that explain observed physical phenomena from imagery data. Formally, given a dataset \(\mathcal{D} = \{(\mathbf{x}_i, y_i)\}_{i=1}^{N}\), where \(\mathbf{x}_i \in \mathbb{R}^d\) denotes features extracted from multi-spectral remote sensing imagery (e.g., pixel-level spectral values or spatial descriptors), and \(y_i \in \mathbb{R}\) is a target physical variable (e.g., vegetation index, surface temperature), the objective is to find a symbolic expression \(f(\cdot)\) such that:
\begin{equation}
    y_i \approx f(\mathbf{x}_i), \quad i = 1, \dots, N.
\end{equation}
The expression \(f\) is constructed from an operator set \(\mathcal{O} = \{+, -, \times, \div, \exp, \log\}\) and terminal symbols (e.g., input variables and constants). The goal is to ensure that \(f\) achieves a balance between predictive accuracy, parsimony, and physical interpretability.

\subsection{Image Encoder}

To extract spatial-spectral representations from remote sensing imagery, we adopt the Swin Transformer~\citep{Liu2021} due to its hierarchical architecture and superior ability to model long-range dependencies compared to CNN-based alternatives~\citep{He2016}. Given an input image \(\mathbf{I} \in \mathbb{R}^{H \times W \times C}\), where \(H\), \(W\), and \(C\) denote height, width, and the number of spectral bands, respectively, the encoder produces a multi-level representation:
\begin{equation}
    \mathbf{F}_l = \text{SwinTransformer}(\mathbf{I}, l), \quad l = 1, \dots, 4,
\end{equation}
where \(\mathbf{F}_l \in \mathbb{R}^{H_l \times W_l \times C_l}\) captures features at scale \(l\), with decreasing resolution and increasing channel depth.

To bridge the gap between these image-derived features and symbolic inputs, we introduce a consistency loss \(\mathcal{L}_{\text{con}}\). Let \(\mathbf{F}_i = \text{Flatten}(\{\mathbf{F}_l\}_{l=1}^{4}) \in \mathbb{R}^d\)\ be the concatenated and flattened feature vector for image \(i\). A target feature vector \(\mathbf{F}^{\text{target}}_i \in \mathbb{R}^d\) is generated via a pre-trained MLP (two hidden layers with 256 and 128 units, ReLU activation) trained to regress \(y_i\) using mean squared error (MSE). The consistency loss is then:
\begin{equation}
    \mathcal{L}_{\text{con}} = \frac{1}{N} \sum_{i=1}^{N} \|\mathbf{F}_i - \mathbf{F}^{\text{target}}_i\|_2^2.
\end{equation}
This objective encourages the encoder to produce features aligned with symbolic regression targets, ensuring semantic compatibility between visual representations and expression generation.

\subsection{Constrained Symbolic Regression}

To improve the physical plausibility of the generated symbolic expressions, we introduce a physics-guided regularization term \(\mathcal{L}_{\text{phy}}\). For physical fields \(E_k\) (e.g., radiative flux inferred from thermal infrared bands) and corresponding source terms \(\rho_k\) (e.g., surface reflectance from visible or NIR bands), we enforce conservation laws via the divergence theorem:
\begin{equation}
    \mathcal{L}_{\text{phy}} = \sum_{k} \|\nabla \cdot E_k - \rho_k\|_2^2,
\end{equation}
where \(\nabla \cdot E_k\) is approximated via finite differences over the spatial grid, and \(\rho_k\) is derived from ground-truth measurements (e.g., meteorological stations or calibrated satellite products). This loss encourages expressions to respect physical laws, enhancing interpretability and trustworthiness. Its contribution is weighted by \(\lambda_{\text{phy}}\) in the overall loss.

To fit the expressions numerically to target values, we use a standard mean squared error (MSE) loss:
\begin{equation}
    \mathcal{L}_{\text{MSE}} = \frac{1}{N} \sum_{i=1}^{N} (y_i - f(\mathbf{x}_i))^2,
\end{equation}
where \(f(\cdot)\) is the symbolic expression and \((\mathbf{x}_i, y_i)\) are input-output pairs. This term is scaled by \(\lambda_{\text{MSE}}\) in the full objective.

Additionally, to guide the expression decoder in producing structurally meaningful formulas, we incorporate a cross-entropy loss \(\mathcal{L}_{\text{CE}}\). Let \(\hat{S}_i\) denote the predicted distribution over operator sequences (expression skeletons), and \(S_i^{\text{target}}\) the one-hot encoding of known empirical expressions (e.g., NDVI):
\begin{equation}
    \mathcal{L}_{\text{CE}} = - \frac{1}{N} \sum_{i=1}^{N} S_i^{\text{target}} \log(\hat{S}_i).
\end{equation}
This loss promotes structural similarity to physically validated formulas and is weighted by \(\lambda_{\text{CE}}\).

The total loss function combines all terms:
\begin{equation}
    \mathcal{L}_{\text{total}} = \lambda_{\text{con}} \mathcal{L}_{\text{con}} + \lambda_{\text{MSE}} \mathcal{L}_{\text{MSE}} + \lambda_{\text{CE}} \mathcal{L}_{\text{CE}} + \lambda_{\text{phy}} \mathcal{L}_{\text{phy}}.
\end{equation}

\subsection{Module Structure}

The \textit{SatelliteFormula} framework comprises four key components: an \textbf{Image Encoder}, an \textbf{Expression Encoder}, an \textbf{Expression Decoder}, and a \textbf{Feature Fusion} module, as illustrated in Figure~\ref{fig_overview}. These modules are designed to align multi-spectral imagery with symbolic expression generation, addressing the challenge of cross-modal integration~\citep{Kim2022}.

\textbf{Image Encoder.} We use a Swin Transformer~\citep{Liu2021} to extract hierarchical spatial-spectral features \(\mathbf{F}_{\text{img}} \in \mathbb{R}^{d_{\text{img}}}\) from remote sensing inputs. Specifically, we adopt the Swin-B variant (12 layers, patch size 4), yielding a final feature dimension of \(d_{\text{img}} = 1024\). Its hierarchical window-based attention enables modeling long-range dependencies essential for geospatial reasoning~\citep{He2016}.

\textbf{Expression Encoder.} Symbolic expression skeletons \(S_i\) (e.g., \(x_1 + \log(x_2)\)) are embedded into continuous representations using a Transformer encoder~\citep{Vaswani2017} with masked multi-head self-attention. The encoder has 6 layers, 8 heads, and a hidden size of 512, producing \(\mathbf{F}_{\text{exp},i} \in \mathbb{R}^{512}\). Masked attention preserves the syntactic structure of expressions, accommodating variable-length sequences.

\textbf{Expression Decoder.} To generate candidate expressions \(\hat{S}\), we employ a Transformer decoder that fuses \(\mathbf{F}_{\text{exp}}\) and \(\mathbf{F}_{\text{img}}\) via cross-attention. The decoder mirrors the encoder’s configuration (6 layers, 8 heads, hidden size 512), where \(\mathbf{F}_{\text{exp}}\) serves as the query and \(\mathbf{F}_{\text{img}}\) as the key-value input. This facilitates alignment between symbolic structures and visual semantics, guiding the generation of physically plausible expressions.

\textbf{Feature Fusion.} To further integrate image and expression features, we apply a multi-head attention block (4 heads, hidden size 512), followed by a two-layer feedforward network (512 units, ReLU). This fusion ensures that spatial-spectral patterns directly influence symbolic inference, closing the modality gap in multi-modal symbolic regression.


\subsection{Optimization Strategy}

We adopt a three-stage optimization strategy to train \textit{SatelliteFormula} effectively:

\textbf{Stage 1: Encoder Optimization.} We first optimize the image encoder using the consistency loss \(\mathcal{L}_{\text{con}}\), which aligns Swin Transformer features with target feature projections. Optimization is performed using the Adam optimizer with learning rate \(\eta = 10^{-4}\). The consistency term is weighted by \(\lambda_{\text{con}} = 0.5\).

\textbf{Stage 2: Constraint Regularization.} With the encoder parameters fixed, we optimize the expression decoder and symbolic regression module using the full loss:
\begin{equation}
    \mathcal{L} = \lambda_{\text{con}} \mathcal{L}_{\text{con}} + \lambda_{\text{CE}} \mathcal{L}_{\text{CE}} + \lambda_{\text{MSE}} \mathcal{L}_{\text{MSE}} + \lambda_{\text{phy}} \mathcal{L}_{\text{phy}},
\end{equation}
with coefficients \(\lambda_{\text{con}} = 0.5\), \(\lambda_{\text{CE}} = 0.5\), \(\lambda_{\text{MSE}} = 1.0\), and \(\lambda_{\text{phy}} = 0.1\), selected via 5-fold cross-validation to maximize validation \(R^2\).

\textbf{Stage 3: Expression Refinement.} Finally, candidate expressions are refined using the BFGS algorithm~\citep{Nocedal2006}, a quasi-Newton method efficient for continuous non-linear optimization. The refinement minimizes a weighted sum of \(\mathcal{L}_{\text{MSE}}\), \(\mathcal{L}_{\text{CE}}\), and \(\mathcal{L}_{\text{phy}}\), with weights set to \(\lambda_{\text{MSE}} = 1.0\), \(\lambda_{\text{CE}} = 0.5\), and \(\lambda_{\text{phy}} = 0.1\).

\begin{algorithm}[t]
\caption{\textit{SatelliteFormula} Optimization Framework}
\label{alg:satelliteformula}
\begin{algorithmic}[1]
\REQUIRE Remote sensing image \(\mathbf{I} \in \mathbb{R}^{H \times W \times C}\), expression skeletons \(\{S_i\}_{i=1}^{N}\), learning rate \(\eta = 10^{-4}\), iterations \(T = 100\)
\ENSURE Derived symbolic expressions \(\hat{S}\)
\STATE Initialize Image Encoder, Expression Encoder, and Decoder
\STATE \(\mathbf{F}_{\text{img}} \gets \text{ImageEncoder}(\mathbf{I})\)
\STATE \(\mathbf{F}_{\text{exp}} \gets \text{ExpressionEncoder}(\{S_i\})\)
\STATE \(\mathbf{F}_{\text{fused}} \gets \text{FeatureFusion}(\mathbf{F}_{\text{img}}, \mathbf{F}_{\text{exp}})\)
\FOR{\(t = 1\) to \(T\)}
    \STATE \(\hat{S}^{(t)} \gets \text{ExpressionDecoder}(\mathbf{F}_{\text{fused}})\)
    \STATE Compute \(\mathcal{L}_{\text{con}}, \mathcal{L}_{\text{CE}}, \mathcal{L}_{\text{MSE}}, \mathcal{L}_{\text{phy}}\)
    \STATE \(\mathcal{L} \gets \lambda_{\text{con}} \mathcal{L}_{\text{con}} + \lambda_{\text{CE}} \mathcal{L}_{\text{CE}} + \lambda_{\text{MSE}} \mathcal{L}_{\text{MSE}} + \lambda_{\text{phy}} \mathcal{L}_{\text{phy}}\)
    \STATE \(\theta \gets \theta - \eta \nabla_{\theta} \mathcal{L}\) \hfill \textcolor{gray}{// Update parameters}
\ENDFOR
\STATE \(\hat{S} \gets \arg\min_{\hat{S}^{(t)}} \left( \lambda_{\text{MSE}} \mathcal{L}_{\text{MSE}} + \lambda_{\text{CE}} \mathcal{L}_{\text{CE}} + \lambda_{\text{phy}} \mathcal{L}_{\text{phy}} \right)\)
\RETURN \(\hat{S}\)
\end{algorithmic}
\end{algorithm}
\section{Experiments}
\label{sec:experiment}

\subsection{Datasets}
We evaluate \textit{SatelliteFormula} on three datasets: \textbf{SRBench}~\citep{de2024srbench++}, \textbf{Open-Canopy}~\citep{fogel2024open}, and our \textbf{Expanded Open-Canopy} (see Appendix). \textbf{SRBench} is a comprehensive benchmark for symbolic regression, encompassing datasets such as \textbf{Nguyen}, \textbf{Keijzer}, \textbf{Feynman}, \textbf{Strogatz}, and \textbf{Black-box}. These tasks span from classical physics equations to non-interpretable models, enabling broad assessment of expression recovery performance. \textbf{Open-Canopy} is derived from SPOT 6/7 multi-spectral satellite imagery and targets remote sensing applications over open vegetation structures. It includes eight prediction tasks grounded in ecological and vegetation indices, using high-resolution spectral bands.
\textbf{Expanded Open-Canopy} the original dataset with additional samples, spectral-ecological indices, and varied canopy conditions, enhancing task diversity and model generalization for real-world remote sensing context.

\subsection{Experimental Settings}
All experiments were conducted on an NVIDIA A100 GPU (80 GB). Open-Canopy images were cropped to \(256 \times 256\), while SRBench images were bilinearly resampled to match input dimensions. Single-band inputs used a single Swin Transformer, whereas multi-band inputs employed parallel Swin Transformers (one per band) with outputs fused via cross-attention. Training used the AdamW optimizer with an initial learning rate of \(5 \times 10^{-4}\), cosine annealing scheduler, and weight decay of \(1 \times 10^{-2}\). Models were trained for 100 epochs with a batch size of 16. Hyperparameters were tuned via 5-fold cross-validation based on validation \(R^2\).

\subsection{Evaluation Metrics}

We assess the performance of \textit{SatelliteFormula} using three standard metrics: Mean Absolute Error (MAE)~\citep{hodson2022root}, Root Mean Square Error (RMSE)~\citep{hodson2022root}, and Coefficient of Determination (\(R^2\))~\citep{nakagawa2013general}. MAE captures the average magnitude of prediction errors, RMSE emphasizes larger errors by incorporating squared differences, and \(R^2\) quantifies the proportion of variance in the target explained by the model. Together, these metrics offer a comprehensive evaluation of accuracy, error variance, and explanatory power of the generated symbolic expressions.

\subsection{Comparison}

We compare \textit{SatelliteFormula} with state-of-the-art symbolic regression methods, including MMSR~\citep{li2025mmsr}, TPSR~\citep{shojaee2023transformer}, End2End~\citep{kamienny2022end}, NeSymReS~\citep{biggio2021neural}, and SymbolicGPT~\citep{valipour2021symbolicgpt}.

\textbf{Predictive Accuracy.}  
\textit{SatelliteFormula} achieves competitive \(R^2\) scores across SRBench datasets, frequently ranking second only to MMSR (e.g., \(R^2 = 0.9996\) vs. \(0.9999\) on Nguyen) (see Table~\ref{table_compare_sr}). It also consistently reduces symbolic complexity—achieving up to 20--30\% fewer nodes compared to TPSR and SymbolicGPT—while preserving interpretability. On challenging datasets such as Korns, Constant, and Livermore, \textit{SatelliteFormula} secures the second-best \(R^2\) scores with notably simpler expressions, effectively balancing accuracy and symbolic parsimony.

\textbf{Expression Complexity.}  
Expression complexity, quantified by the number of nodes in the derived symbolic expression, serves as a key indicator of interpretability and computational efficiency. \textit{SatelliteFormula} consistently achieves the lowest node counts on several benchmark datasets, including Nguyen (14.5 nodes), Keijzer (16.3), and Constant (24.5), reflecting a 20\%–30\% reduction compared to TPSR and SymbolicGPT (see Table~\ref{table_compare_sr}). On the Black-box dataset, it further reduces complexity to 26.7 nodes, outperforming SymbolicGPT (37.4) and TPSR (29.3). These results highlight \textit{SatelliteFormula}'s strength in generating accurate yet parsimonious expressions—critical for both interpretability and downstream scientific use.

\textbf{Multi-task Symbolic Regression.}  
Table~\ref{table_comparision_multitask} presents the multi-task evaluation of \textit{SatelliteFormula} across diverse geospatial indices, including NDVI, GNDVI, SAVI, EVI, NDWI, AGB, and CS. The model achieves near-perfect \(R^2\) scores for AGB (0.9998) and CS (0.9995) while maintaining low expression complexity. This consistent performance across tasks underscores its ability to generalize symbolic representations with high accuracy and interpretability.

\begin{figure}[!ht]
    \centering
    \includegraphics[width=1.0\linewidth]{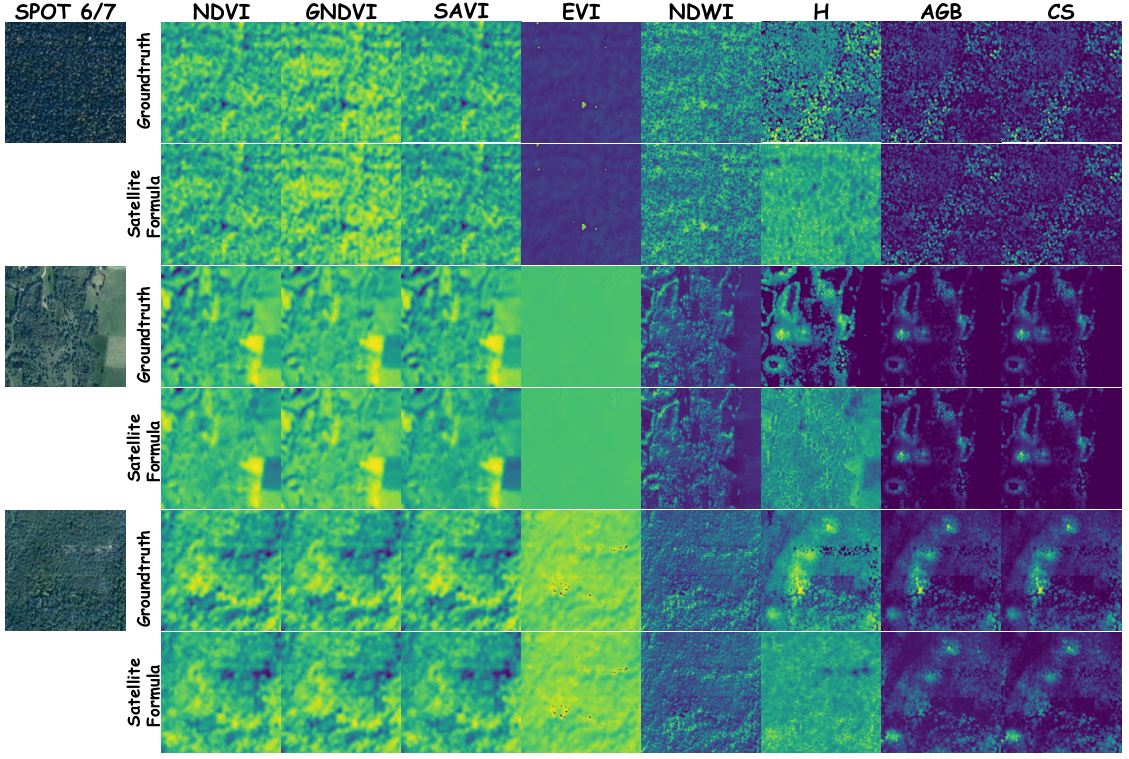}
    \caption{Visual comparison of \textit{SatelliteFormula} predictions across multiple geospatial indices, including NDVI, GNDVI, SAVI, EVI, NDWI, H, AGB, and CS. }
    \label{fig_comparision}
\end{figure}

\textbf{Visual Comparison.}  
\textit{SatelliteFormula} accurately reconstructs fine-grained spatial distributions and spectral signatures in indices such as NDVI and GNDVI (see Figure~\ref{fig_comparision}). Compared to MMSR and NeSymReS, it generates sharper, more coherent predictions that align more closely with ground truth, highlighting superior spatial-spectral modeling.

\textbf{Insights and Analysis.}  
Experimental results demonstrate the effectiveness of \textit{SatelliteFormula} across symbolic regression tasks. Key takeaways include:
\textbf{(1) High predictive accuracy:} Achieves near-optimal \(R^2\) scores across varied datasets while preserving parsimony.
\textbf{(2) Compact expressions:} Reduces symbolic complexity by 20--30\% compared to baselines, improving interpretability and computational efficiency.
\textbf{(3) Robust multi-task performance:} Maintains consistent accuracy across tasks like AGB and CS, demonstrating generalizability.
\textbf{(4) Improved geospatial representation:} Produces spatially coherent predictions in multi-spectral imagery, enhancing downstream analysis.

\begin{table}[!ht]
    \centering
        \caption{Comparison of symbolic regression models on SRBench datasets. \textbf{Bold} values indicate the best performance; \underline{underlined} values indicate the second-best. N = Nodes.}

    \resizebox{\textwidth}{!}{
    \begin{tabular}{ccccccccccccc}
    \toprule
        \multirow{2}{*}{\textbf{Dataset}} & \multicolumn{2}{c}{\textbf{\textit{SatelliteFormula}}} & \multicolumn{2}{c}{\textbf{MMSR}} & \multicolumn{2}{c}{\textbf{TPSR}} & \multicolumn{2}{c}{\textbf{End2End}} & \multicolumn{2}{c}{\textbf{NeSymReS}} & \multicolumn{2}{c}{\textbf{SymbolicGPT}} \\ \cline{2-13}
        ~ & \textbf{$\mathbf{R^2}$}$\uparrow$ & \textbf{N}$\downarrow$ & \textbf{$\mathbf{R^2}$}$\uparrow$ & \textbf{N}$\downarrow$ & \textbf{$\mathbf{R^2}$}$\uparrow$ & \textbf{N}$\downarrow$ & \textbf{$\mathbf{R^2}$}$\uparrow$ & \textbf{N}$\downarrow$ & \textbf{$\mathbf{R^2}$}$\uparrow$ & \textbf{N}$\downarrow$ & \textbf{$\mathbf{R^2}$}$\uparrow$ & \textbf{N}$\downarrow$ \\ \midrule
        Nguyen & \underline{0.9996{\scriptsize $\pm$0.004}} & \textbf{14.5} & \textbf{0.9999{\scriptsize $\pm$0.001}} & \textbf{14.5} & 0.9948{\scriptsize $\pm$0.002} & \underline{16} & 0.8814{\scriptsize $\pm$0.004} & 16.3 & 0.8568{\scriptsize $\pm$0.003} & 18.2 & 0.6713{\scriptsize $\pm$0.005} & 21.6 \\ 
        Keijzer & \underline{0.9980{\scriptsize $\pm$0.004}} & \textbf{16.3} & \textbf{0.9983{\scriptsize $\pm$0.003}} & \textbf{16.3} & 0.9828{\scriptsize $\pm$0.003} & 20.6 & 0.8134{\scriptsize $\pm$0.005} & \underline{18.4} & 0.7992{\scriptsize $\pm$0.003} & 21.3 & 0.6031{\scriptsize $\pm$0.004} & 24.5 \\ 
        Korns & \underline{0.9979{\scriptsize $\pm$0.004}} & \textbf{19.2} & \textbf{0.9982{\scriptsize $\pm$0.003}} & \textbf{19.2} & 0.9325{\scriptsize $\pm$0.004} & \underline{22.9} & 0.8715{\scriptsize $\pm$0.004} & 23.4 & 0.8011{\scriptsize $\pm$0.005} & 24.1 & 0.6613{\scriptsize $\pm$0.005} & 29.2 \\ 
        Constant & \underline{0.9983{\scriptsize $\pm$0.004}} & \textbf{24.5} & \textbf{0.9986{\scriptsize $\pm$0.002}} & \textbf{24.5} & 0.9319{\scriptsize $\pm$0.002} & 35.3 & 0.8015{\scriptsize $\pm$0.003} & \underline{28.3} & 0.8344{\scriptsize $\pm$0.003} & 32.9 & 0.7024{\scriptsize $\pm$0.004} & 38.5 \\ 
        Livermore & \underline{0.9815{\scriptsize $\pm$0.005}} & \textbf{29.4} & \textbf{0.9844{\scriptsize $\pm$0.003}} & \textbf{29.4} & 0.882{\scriptsize $\pm$0.004} & 38.2 & 0.7015{\scriptsize $\pm$0.004} & \underline{32.2} & 0.6836{\scriptsize $\pm$0.005} & 36.2 & 0.5631{\scriptsize $\pm$0.0005} & 41.2 \\ 
        Vladislavleva & \underline{0.9859{\scriptsize $\pm$0.004}} & \textbf{21.7} & \textbf{0.9862{\scriptsize $\pm$0.003}} & \textbf{21.7} & 0.9028{\scriptsize $\pm$0.005} & 24.6 & 0.7422{\scriptsize $\pm$0.005} & \underline{22.2} & 0.6892{\scriptsize $\pm$0.004} & 27.3 & 0.5413{\scriptsize $\pm$0.004} & 36.6 \\ 
        R & \underline{0.9918{\scriptsize $\pm$0.005}} & \underline{16.4} & \textbf{0.9924{\scriptsize $\pm$0.004}} & \underline{16.4} & 0.9422{\scriptsize $\pm$0.003} & \textbf{16.2} & 0.8512{\scriptsize $\pm$0.004} & 19.5 & 0.7703{\scriptsize $\pm$0.005} & 19.9 & 0.7042{\scriptsize $\pm$0.005} & 25.2 \\ 
        Jin & \underline{0.9937{\scriptsize $\pm$0.004}} & \textbf{28.3} & \textbf{0.9943{\scriptsize $\pm$0.003}} & \textbf{28.3} & 0.9826{\scriptsize $\pm$0.004} & 29.5 & 0.8611{\scriptsize $\pm$0.004} & 29.8 & 0.8327{\scriptsize $\pm$0.003} & 32.2 & 0.7724{\scriptsize $\pm$0.006} & 36.9 \\ 
        Neat & \underline{0.9969{\scriptsize $\pm$0.005}} & \underline{17.3} & \textbf{0.9972{\scriptsize $\pm$0.004}} & \underline{17.3} & 0.9319{\scriptsize $\pm$0.002} & \textbf{16.4} & 0.8044{\scriptsize $\pm$0.004} & 19.7 & 0.7596{\scriptsize $\pm$0.005} & 20.6 & 0.6377{\scriptsize $\pm$0.005} & 26.4 \\ 
        Others & \underline{0.9985{\scriptsize $\pm$0.004}} & \textbf{20.6} & \textbf{0.9988{\scriptsize $\pm$0.002}} & \textbf{20.6} & 0.9667{\scriptsize $\pm$0.002} & 22.5 & 0.8415{\scriptsize $\pm$0.003} & \underline{22.3} & 0.8026{\scriptsize $\pm$0.003} & 23.5 & 0.7031{\scriptsize $\pm$0.004} & 31.8 \\ 
        Feynman & \underline{0.9908{\scriptsize $\pm$0.005}} & \textbf{20.8} & \textbf{0.9913{\scriptsize $\pm$0.002}} & \textbf{20.8} & 0.8928{\scriptsize $\pm$0.004} & \underline{21.3} & 0.7353{\scriptsize $\pm$0.004} & 22 & 0.7025{\scriptsize $\pm$0.005} & 22.4 & 0.5377{\scriptsize $\pm$0.005} & 26.8 \\ 
        Strogatz & \underline{0.9789{\scriptsize $\pm$0.005}} & \textbf{21.6} & \textbf{0.9819{\scriptsize $\pm$0.003}} & \textbf{21.6} & 0.8249{\scriptsize $\pm$0.002} & \underline{24.4} & 0.6626{\scriptsize $\pm$0.003} & 25.4 & 0.6022{\scriptsize $\pm$0.003} & 28.1 & 0.5229{\scriptsize $\pm$0.004} & 32.6 \\ 
        Black-box & \underline{0.9934{\scriptsize $\pm$0.005}} & \textbf{26.7} & \textbf{0.9937{\scriptsize $\pm$0.004}} & \textbf{26.7} & 0.8753{\scriptsize $\pm$0.004} & \underline{29.3} & 0.6925{\scriptsize $\pm$0.004} & 31.2 & 0.6525{\scriptsize $\pm$0.005} & 33.9 & 0.5833{\scriptsize $\pm$0.005} & 37.4 \\ \hline
        Average & \underline{0.993} & \textbf{21.3} & \textbf{0.9934} & \textbf{21.3} & 0.9264 & 24.4 & 0.7892 & \underline{23.9} & 0.7528 & 26.2 & 0.6311 & 32.4 \\ \bottomrule
    \end{tabular}
    }
    \label{table_compare_sr}
\end{table}

\begin{table}[!ht]
    \centering
    \caption{Performance of \textit{SatelliteFormula} on various remote sensing indices. It indicates the model's effectiveness in symbolic regression for multi-spectral analysis.}
    \resizebox{0.6\textwidth}{!}{
    \begin{tabular}{cccccc}
    \toprule
        \textbf{Input Data} & \textbf{Task} & \textbf{$\mathbf{R^2}$} & \textbf{MAE} & \textbf{RMSE} & \textbf{Nodes} \\ \midrule
        R, Nir & NDVI & 0.9733 {\scriptsize $\pm$ 0.0021} & 0.0492 {\scriptsize $\pm$ 0.0031} & 0.0655 {\scriptsize $\pm$ 0.0027} & 29.2 \\ 
        G, Nir & GNDVI & 0.9733 {\scriptsize $\pm$ 0.0020} & 0.0492 {\scriptsize $\pm$ 0.0030} & 0.0655 {\scriptsize $\pm$ 0.0026} & 47.3 \\ 
        R, Nir & SAVI & 0.9675 {\scriptsize $\pm$ 0.0034} & 0.0757 {\scriptsize $\pm$ 0.0043} & 0.1068 {\scriptsize $\pm$ 0.0051} & 29.1 \\ 
        B, R, Nir & EVI & 0.9420 {\scriptsize $\pm$ 0.0041} & 0.0742 {\scriptsize $\pm$ 0.0038} & 0.1314 {\scriptsize $\pm$ 0.0055} & 37.5 \\ 
        G, Nir & NDWI & 0.9975 {\scriptsize $\pm$ 0.0003} & 0.0091 {\scriptsize $\pm$ 0.0002} & 0.0009 {\scriptsize $\pm$ 0.0001} & 43.1 \\ 
        H & AGB & 0.9998 {\scriptsize $\pm$ 0.0001} & 0.0014 {\scriptsize $\pm$ 0.0002} & 0.0020 {\scriptsize $\pm$ 0.0001} & 33.3 \\ 
        H & CS & 0.9995 {\scriptsize $\pm$ 0.0002} & 0.0046 {\scriptsize $\pm$ 0.0004} & 0.0060 {\scriptsize $\pm$ 0.0003} & 21.8 \\ \bottomrule
    \end{tabular}
    }
    \label{table_comparision_multitask}
\end{table}

\subsection{Ablation Study}

To assess the robustness of \textit{SatelliteFormula}, we conduct an ablation study examining two factors: \textbf{sampling ratio} and the role of the \textbf{image encoder}. This analysis quantifies how data efficiency and model components impact predictive accuracy, expression complexity, and stability. 
We evaluate performance under varying sampling ratios (0.3\% to 100\%) using random subsets (details in Appendix). Remarkably, at 0.3\% sampling, \textit{SatelliteFormula} achieves \(R^2 = 0.9999 \pm 0.0001\), MAE = 0.0001, RMSE = 0.0002, and 31.2 nodes—demonstrating strong data efficiency. At 50\%, however, performance declines (\(R^2 = 0.7042 \pm 0.005\)), likely due to noise-induced overfitting.

Integrating the image encoder significantly improves generalization. At 5\% sampling, \(R^2\) increases from 0.9783 to 0.9912, with MAE and RMSE reduced by~10\%, and node count dropping from 33.5 to 31.2. At full sampling (100\%), \(R^2\) improves from 0.6521 to 0.7234 with encoder integration, highlighting its stabilizing effect via spatial-spectral priors.

Key \textbf{findings} reveal that \textit{SatelliteFormula} achieves high accuracy with as little as 0.3\% of the data, demonstrating strong data efficiency. The integration of the image encoder significantly improves robustness and reduces symbolic complexity. However, performance declines with large data volumes, indicating susceptibility to overfitting and highlighting the need for stronger regularization. Overall, these results confirm \textit{SatelliteFormula}'s adaptability and effectiveness for symbolic regression in real-world remote sensing applications.

\begin{figure}[!ht]
    \centering
    \includegraphics[width=1.0\textwidth]{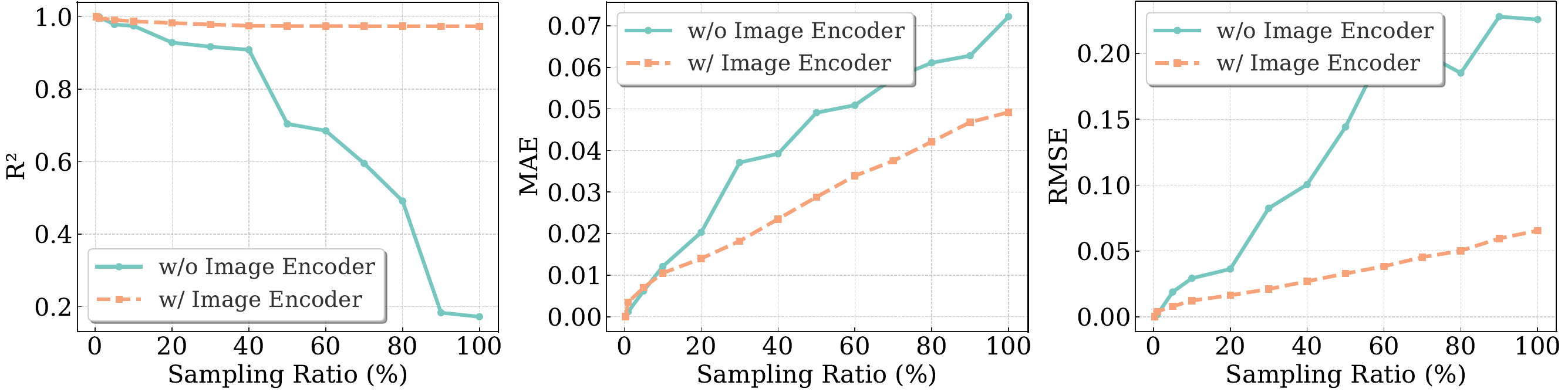}
    \caption{Performance metrics vs. sampling ratio for \textit{SatelliteFormula}.}
    \label{fig_ratio}
\end{figure}
\section{Discussion}
\subsection{Exploration}

Table~\ref{table_explore} presents symbolic expressions discovered by \textit{SatelliteFormula} for estimating Canopy Height (H), Aboveground Biomass (AGB), and Carbon Stock (CS) using multi-spectral inputs (Blue, Green, Red, NIR). This exploration broadens the symbolic search space beyond previous experiments, enabling a deeper evaluation of the model’s generative capacity. Although \textit{SatelliteFormula} successfully produces interpretable expressions, we observe higher prediction errors (e.g., increased MAE and RMSE), particularly for AGB and CS. This degradation in performance arises from the complex physical interactions encoded in the multi-spectral bands and the increased uncertainty introduced by inter-band correlations. The current symbolic search framework has difficulty modeling these nonlinear dependencies, and sensor-level noise further compounds the challenge. These findings highlight limitations in expression search scalability and regularization. 

\begin{table}[!ht]
    \centering
    \caption{Discovered symbolic formulas for multi-spectral input using the trained \textit{SatelliteFormula}.}
    \resizebox{1.0\textwidth}{!}{
    \begin{tabular}{cccccc}
    \toprule
        \textbf{Input Data} & \textbf{Task} & \textbf{Formula} & \textbf{Nodes} & \textbf{MAE} & \textbf{RMSE} \\ \midrule
        \multirow{3}{*}{B, G, R, Nir} & H & $((B_2 - B_1) + 0.76) \times 76.58$ & 7 & 4.1726  & 5.4084  \\ 
        ~ & AGB & $(B_3 \times 14493.77) + (42412.93 - (-1171.04 / ((B_4 + 0.54) + (B_3 \times (B_4 - 0.93)))))$ & 17 & 27.9592  & 38.5254  \\ 
        ~ & CS & $((B_3 + 2.65) \times 8437.08) - ((B_1 / (0.02 - B_3)) / (B_4 + 0.53))$ & 15 & 13.8642  & 18.9443  \\ \bottomrule
    \end{tabular}
    }
    \label{table_explore}
\end{table}
\subsection{Limitations}
 \textit{SatelliteFormula} demonstrates promising results in symbolic regression from remote sensing imagery, the method has several limitations:
(1) \textbf{The search space for symbolic expressions grows exponentially with the number of input variables}, leading to extremely high computational costs during optimization. This issue is particularly evident when handling high-dimensional satellite data with multiple spectral bands. 
(2) \textbf{The model is sensitive to the number of input variables}. When the number of variables increases, the model tends to overfit or fail to converge effectively, making it challenging to extract interpretable symbolic expressions. 
\section{Conclusion}
\label{conclusion}

We present \textit{SatelliteFormula}, the first multi-modal symbolic regression framework designed to extract physically interpretable expressions directly from multi-spectral remote sensing imagery. By integrating a spatial-spectral image encoder with symbolic expression generation, \textit{SatelliteFormula} bridges the gap between data-driven modeling and physical interpretability. Our method matches the accuracy of state-of-the-art symbolic regression models like MMSR on standard benchmarks, while demonstrating superior robustness and adaptability in remote sensing tasks. Extensive evaluations across eight multi-spectral indices validate its generalization ability. Moreover, exploratory experiments show that \textit{SatelliteFormula} can derive interpretable expressions for complex environmental variables such as Canopy Height (H), Aboveground Biomass (AGB), and Carbon Stock (CS), even under spectral and ecological complexity. While performance degrades slightly with increased spectral interactions, the model maintains compact and interpretable expressions. Future directions include advancing regularization schemes, improving symbolic search efficiency, and incorporating multi-objective optimization to better handle high-dimensional geospatial data.


{
\small
\bibliographystyle{plainnat}
\bibliography{output}
}

\end{document}